%% file: main.tex
\documentclass[letterpaper, 10 pt, conference]{ieeeconf/ieeeconf_latest}  
\input{macros}

\usepackage[sort,compress]{cite}
\newcommand{\NOTE}[1]{\textcolor{red}{\bf NOTE: {#1}}}

\usepackage{hyperref}
\usepackage{mathtools}
\usepackage{xcolor}
\usepackage{bm}
\usepackage{gensymb}
\usepackage{amsmath}
\usepackage{graphicx}
\usepackage{array,multirow}
\usepackage{float} 
\usepackage[misc]{ifsym} 
\usepackage[paper=letterpaper,margin=0.75in]{geometry} 

\IEEEoverridecommandlockouts
\title{\LARGE \bf Defensive Escort Teams via Multi-Agent Deep Reinforcement Learning}

\author{Arpit Garg$^{1}$, Yazied A. Hasan$^{1}$, Adam Ya{\~n}ez$^{1}$ and Lydia Tapia$^{1}$
\thanks{*This work is partially supported by the National Science Foundation (NSF) under Grant Numbers IIS-1528047 and IIS-1553266.  This work is also partially supported by the Air Force Research Laboratory (AFRL) under agreement number FA9453-18-2-0022.  Also, partial support was provided by the Army Research Laboratory (ARL) and was accomplished under Cooperative Agreement Number W911NF-19-2-0215. Any opinions, findings,  conclusions, recommendations, or views  contained in this document are those of the authors and should not be interpreted as representing the official policies, either expressed or implied, of the NSF, AFRL, ARL or U.S. Government. The U.S. Government is authorized to reproduce and distribute reprints for Government purposes notwithstanding any copyright notation herein. }
\thanks{$^{1}$Department of Computer Science, University of New Mexico, MSC01 11301 University of New Mexico, Albuquerque, NM 87131, USA
        {\tt\small kiralobo@cs.unm.edu, yhasan@unm.edu, ayanez2@unm.edu, and (corresponding) tapia@cs.unm.edu}}%
}

\begin{document}

\maketitle
\thispagestyle{empty}
\pagestyle{empty}

\begin{abstract}
Coordinated defensive escorts can aid a navigating payload by positioning themselves in order to maintain the safety of the payload from obstacles.  In this paper, we present a novel, end-to-end solution for coordinating an escort team for protecting high-value payloads. Our solution employs deep reinforcement learning (RL) in order to train a team of escorts to maintain payload safety while navigating alongside the payload.  This is done in a distributed fashion, relying only on limited range positional information of other escorts, the payload, and the obstacles. When compared to a state-of-art algorithm for obstacle avoidance, our solution with a single escort increases navigation success up to 31\%.  Additionally, escort teams increase success rate by up to 75\% percent over escorts in static formations.  We also show that this learned solution is general to several adaptations in the scenario including:  a changing number of escorts in the team, changing obstacle density, and changes in payload conformation. 
Video: \href{https://youtu.be/SoYesKti4VA}{https://youtu.be/SoYesKti4VA}.
\end{abstract}

\section{Introduction}

 
 Successful navigation in crowded scenarios often requires assuming a non-zero collision probability between the agent and stochastic obstacles \cite{chiang2017safety}.   This required assumption of risk is potentially frightening given the value of cargo that modern autonomous agents will be transporting, e.g., human life.  
In many real-world scenarios, humans employ escorts for enhanced safety during  high-consequence navigation, e.g., a parent with a child, presidential security, or military convoys in dangerous environments.  For example, the US Army employs a tactical convoy 
to move a payload, personnel and/or cargo, via a group of ground vehicles 
to or from a target destination. Some of the vehicles in the convoy act as coordinated escorts to prevent traffic from overtaking the convoy, dispersing crowds, or establishing a secure perimeter (cordon area) that is essential to the safety of the soldiers.
Robotic escorts have also been employed to aid navigation, such as a suitcase that sounds audible warnings upon expected collision \cite{kayukawa-bbeep}, a robot guide dog for visually-impaired humans \cite{tachi1984guide}, or an autonomous shopping cart \cite{gharpure2008robot}.  Even tactical teams have employed escorting agents providing reconnaissance, e.g., video previews of the environment, \cite{bethel2012discoveries,lalejini2014evaluation}.  However, in all these scenarios, the escort's feedback is used to determine if the payload's navigation route should continue or change, rather than to provide safe passage.  

\begin{figure}[tb]
    \centering
    \includegraphics[scale=0.13]{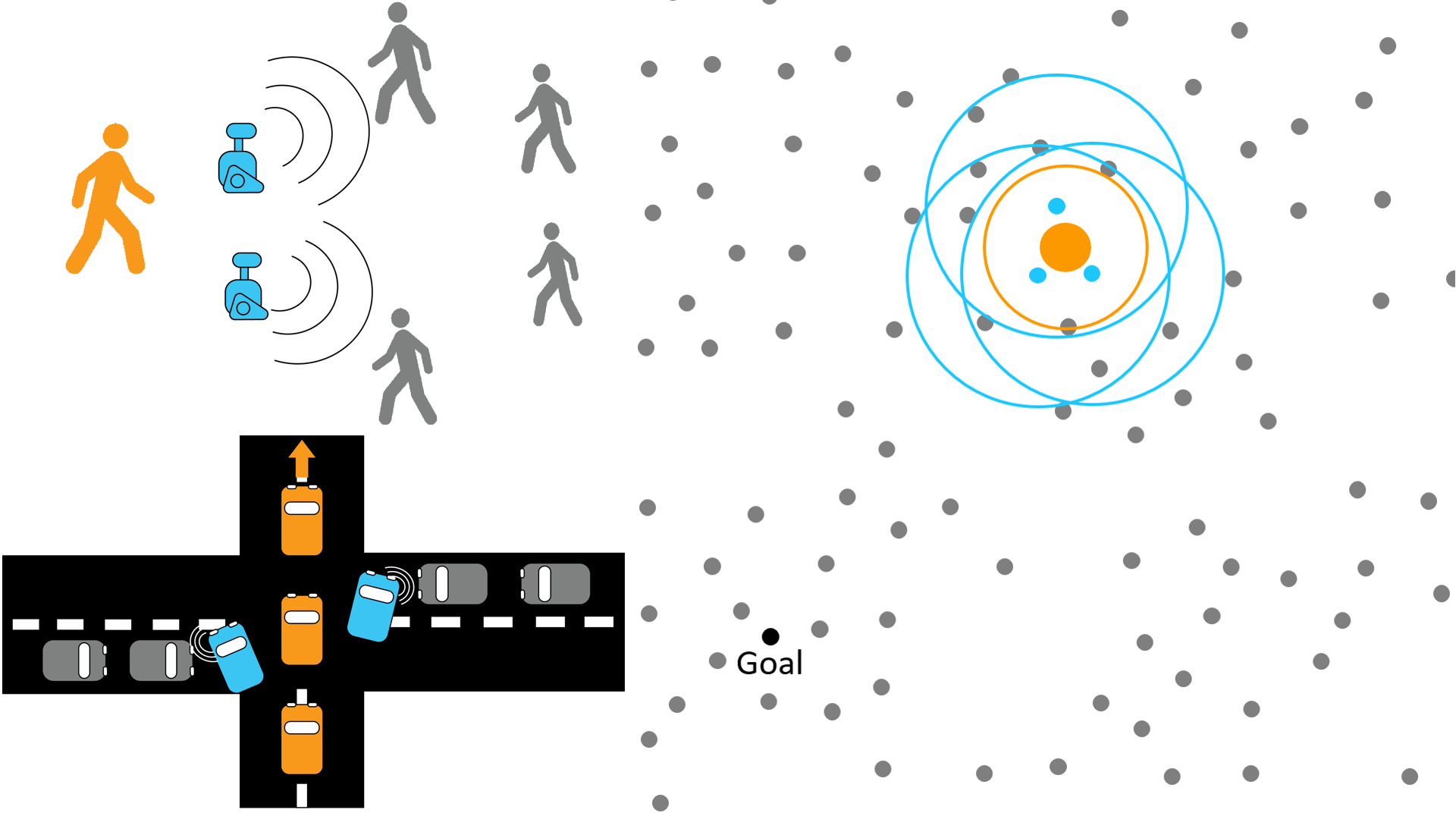}
    \caption{\scriptsize A team of defensive escorts (blue) to protect a payload (orange). (Left) Examples of escort scenarios including an escort team clearing the path for a human through a crowd or an escort for a vehicle caravan. (Right) Our experimental setup demonstrates the payload (large orange dot) navigating to the goal (black dot) with coordinated escorts (blue dots) interacting with obstacles (grey dots).
    The blue outlined circles indicate the sensor radius of the escorts.  The orange outlined circle indicates a cordon safety area around the payload.}
    \label{fig:trainingEnvironment}
\end{figure}

For payloads with high collision consequences, navigation hindrance may not be an option.  Coordinated escorts must clear the route for the payload in order to continue safe navigation. Therefore, we present an autonomous fully distributed solution to this navigation scenario where a team of escorts learn to defend the payload by interacting with obstacles that are expected to collide with the payload, e.g., by pushing obstacles away (see Fig.~\ref{fig:trainingEnvironment}). 
Our end-to-end solution takes  partial observations of the environment and no other explicit information. 
The escorts learn to automatically position themselves around the payload and adapt to obstacles with stochastic and interacting motions.
This is achieved without explicit communication between escorts or between the escort and the payload via deep reinforcement learning.

Our deep RL solution provides enhanced safety of the payload along a fixed navigation route as compared to agile maneuvering of the payload using a state-of-the-art obstacle avoidance algorithm \cite{arpit-RLvsSR}.  We also show that a dynamic and responsive escort team outperforms fixed formations of escorts.  Additionally, our learned solution is flexible to changes in scenario including adapting to a changing conformation of the payload, dynamic changes in the numbers of escorts in the team, and obstacle density. Specifically, we show that a single deep RL escort
increases the success rate by up to 31\% 
compared to a state-of-the-art agile payload navigation algorithm. Additionally, escort teams that provide dynamic interception increase the success rate by up to 75\% 
when compared to escorts positioned in static formations around the payload.

This paper provides several novel contributions including: a definition of the Payload Protection Problem that identifies the roles and collaboration between the payload and the escort team in a crowded environment, a deep RL solution to the Payload Protection Problem providing enhanced safety for the payload and cordon area around the payload, and adaptation of the deep RL solution to different problem variations both with and without retraining.  Demonstrations of the method are provided in the accompanying video.

\excise{
\section{Adam's Apple :D}
Defender teams can be used as part of armed forces operations, such as tactical convoys. The US Army defines a tactical convoy as a deliberately planned combat operation to move personnel and/or cargo via a group of ground transportation assets in a secure manner to or from a target destination under the control of a single commander in a permissive, uncertain, or hostile environment. The presence of civilians can complicate an operation. Defender teams can help alleviate some of the complexity by taking on roles usually allotted to soldiers. During a convoy, defender teams can help provide security by controlling civilian traffic about the convoy. This includes preventing traffic from overtaking the convoy on a highway or at intersections. Another example is for dispersing crowds while in motion or dismounted. A defender team can help maintain 360-degree security by cordoning off an area from civilians. This would establish a secure perimeter essential to the safety of the soldiers.
}

\section{Related Work}

Robots have been used to provide human navigational assistance in a variety of scenarios.   For example, robots have been developed to help a visually impaired human navigate via a guide robot dog \cite{tachi1984guide} and an autonomous shopping cart \cite{gharpure2008robot}.  Recently, a non-mobile robot suitcase system was developed for assisting visually impaired human navigation \cite{kayukawa-bbeep}. The suitcase uses computer vision to predict future collisions and generates an audible warning to warn the incoming, often distracted, pedestrians.
 In emergency navigation scenarios,  audio navigational cues were also found to be helpful for mobile robots \cite{shell2005insights,robinette2013building}.
Tactical team assistance was provided by multiple mobile robots providing reconnaissance, e.g., video previews of the environment, \cite{bethel2012discoveries,lalejini2014evaluation}.
In another example, firefighters were guided in extremely low visibility conditions by a group of mobile robots \cite{penders2010guardians}.
In all these navigational assistance scenarios, the robots do not enhance navigational safety by intercepting potential risks or actively manipulating the environment.

Robotic agents that exert influence over the environment in order to achieve goals have been investigated extensively for group control problems, i.e., navigation or control of a group of agents using one or several robots \cite{vo2009behavior}.
Common applications of group control includes crowd evacuation planning \cite{brenner2005simulating}, pollution control \cite{fingas2012basics}, and animal herding \cite{strombom2014solving}.
Typical approaches for this highly under-actuated system includes sampling-based methods such as RRT or PRM \cite{vo2009behavior}, social force-based methods \cite{martinez2005crowding}, and nature-inspired rule-based heuristics \cite{strombom2014solving}.  In contrast to these approaches, our learned solution is dynamic to several problem changes and interacts with both the payload and the obstacles in the environment to enhance transport safety.

The perimeter surveillance problem is a subset of the group control problem, where multiple robots circulate around a moving region in order to create a virtual fence.
The virtual fence can be used to control or protect the entities inside from colliding with static obstacles \cite{jahn2017distributed,saldana2016dynamic}. 
To achieve such goals, the parameterized shape of the fence is often first planned by a sampling-based motion planner,  and then reactive controllers are used on each robot to follow the shape of the fence \cite{jahn2017distributed,saldana2016dynamic}.
Our autonomous escort team addresses a similar problem, however, our escorts operate in a dynamic environment and use the partial LiDAR observation rather than assuming the full knowledge of obstacles. In addition, the escorts are not restricted to orbiting motion and adapt to changes in team objective, number of escorts, and obstacle motion uncertainties.

Shepherding is another subset of the group control problem  \cite{lien-shepherd-planning, lien-shepherding-behavior}. 
Shepherding involves one or many shepherd agents that attempt to control (e.g., via exerting repulsive forces) the motions of herd agents to achieve goals such as moving the herd to a specified location.
To achieve such goals, a simple, dog behavior-inspired shepherding heuristic based on the herd center of mass and the furthest herd was developed in \cite{strombom2014solving}.
In comparison, although our autonomous escort team also influences obstacles via repulsive forces, our objective is navigation safety in a dynamic environment rather than transport of multiple shepherd agents.

\excise{
Another type of group control problem do not use mobile robots. 
Instead, an assistive suitcase system, BBeep, was developed for visually impaired people \cite{kayukawa-bbeep}. BBeep uses computer vision to predict future collisions between the user and nearby pedestrians. If a collision is predicted, it generates an audible sound to warn the incoming, often distracted pedestrians to avoid collision.
In contrast, not only do our escort agents predict potential obstacle-payload collisions, they also actively re-position themselves to protect the payload.}

The territory guarding problem is a differential game where the invader agents attempt to breach a cordon area while the guarding agents stop the invaders by capturing them as far from the cordon area as possible \cite{isaac-differential-games, lee-differential-fuzzy, lee-guarding-territory-two-invaders}. When imprecise information about the location of the invader is known, an optimized fuzzy controller could be used to locate invader's position \cite{hsia-guarding-territory}.
Fuzzy RL techniques such as fuzzy Q-learning \cite{glorennec-fuzzy-q} and fuzzy actor-critic \cite{Wang-fuzzy-actor-critic} have been used to approximate the optimal solution for both the invader and guarding agents \cite{lu-multi-agent-RL, raslan-learning-invader, analikwu-multi-agent-guarding-territory}. However, these methods also require information about the state of the environment, e.g., the strategy of the guards or invaders and the position of the cordon area, in order to find optimal strategies.
In our work, the autonomous escorts do not capture the invaders but intercept them by applying forces on them to maintain payload safety. In addition, our escort team defends the payload from up to 90 interacting obstacles, an order of magnitude more than the number of invaders considered in existing solutions for the territory guarding problem.

Deep RL has recently shown great success on highly dynamic navigation tasks.  Some methods combine long-range planning with highly adaptable short-range deep RL solutions that continually replan in order to navigate collision free \cite{faust-prm-rl, chiang-rl-rrt}. Some of our prior work presented a deep RL solution for navigating in dynamic environments and compared the learned collision probabilities against a formal and complete method \cite{arpit-RLvsSR}.  Other navigation-based problems like Pursuit-Evasion 
and Waterworld 
have been previously studied by extending deep RL algorithms 
to cooperative multi-agent systems \cite{gupta-coop-multi-agent} that do not use any explicit communication. 
Although, these solutions involve dynamic obstacle avoidance and learning cooperative navigation, the navigation objectives do not involve enhancing safety by escorting a moving payload.

\begin{figure*}[ht]
  \includegraphics[keepaspectratio=true,scale=0.35]{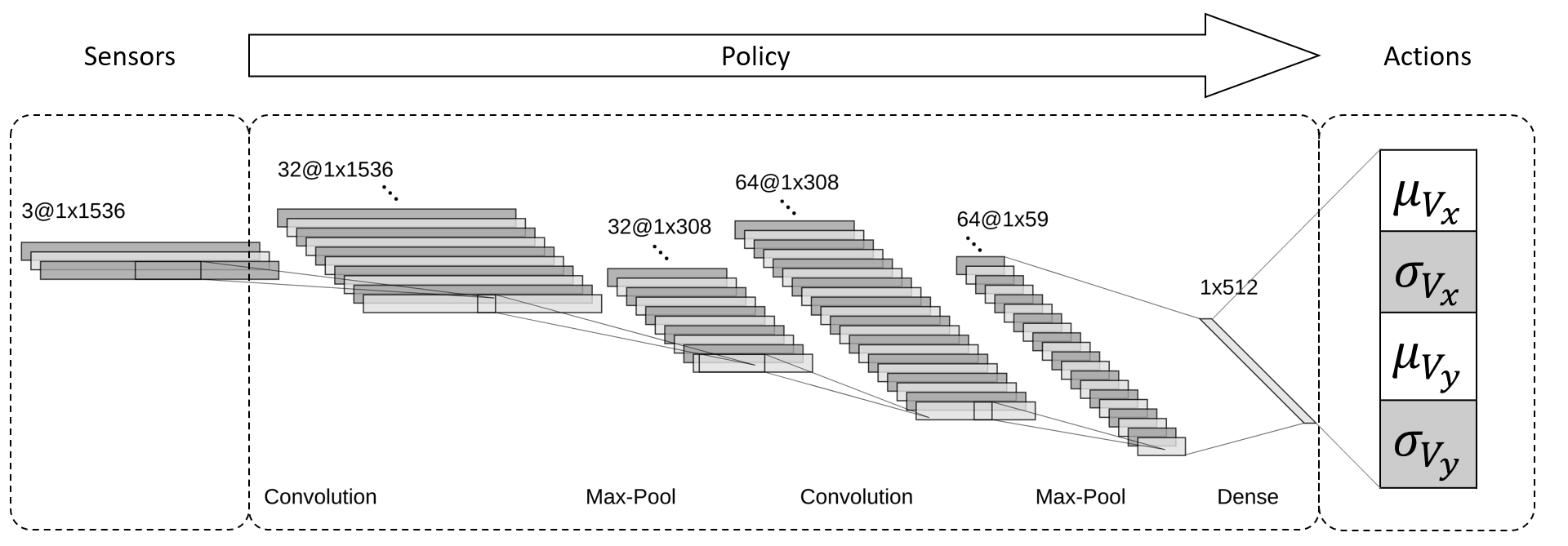}
    \caption{\scriptsize Neural network architecture. The network takes in the sensor information from each type of sensed object:  payload, obstacles, and other escorts, and outputs a diagonal Gaussian distribution from which continuous actions are sampled. The network consists of 2 sets of alternating convolutional and max-Pooling layers followed by a flattened dense layer.}
    \label{fig:NN}
\end{figure*}

\section{Problem Formulation}
\label{sec:problem}


The {\em Payload Protection Task} aims to find an escort policy 
parameterized by $\mathbf{\theta}$ that maps observations 
to robot action 
while maximizing payload safety.  Safety is enforced by minimal probability of collision while navigating to the goal.  Collision events, $\mathbb{C}$, are impacts that involve the payload and/or an escort.  
Additionally, it is often critical to defend a zone around the payload by minimizing the probability of any obstacles entering a cordon area around the payload and the probability of collision, $\mathbb{B}$.    
This translates to 
\begin{align}
    \pi^\text{PP}_{\mathbf{\theta}} = \arg \min_{\bm{\mathbf{\theta}}} \mathbb{P} \big( \mathbb{B} \cup
    \mathbb{C}\big). \label{eq:problem}
\end{align}
This task is critical for applications with high collision consequence. 

For the solution of $\pi^\text{PP}_{\mathbf{\theta}}$, we consider a fully cooperative and partially observable multi-agent task in which each controllable agent, i.e., an escort in the team, receives a private observation that is correlated with the true state of the environment. The escorts have no explicit communication and must learn cooperative behaviour only from their partial observations.
Formally, the problem can be formulated as a multi-agent extension of Partially Observable Markov Decision Process (POMDP), given as a tuple $(S, \mathcal{A}, \mathcal{O}, D, R, \mathcal{T}, \rho, N, \gamma)$, where agents are the escorts. At each step, for a given state $s \in S$, the escort $i \in \{1, ..., N\}$, receives a partial observation $o_i \in \mathcal{O}$, determined by the conditional observation probability $\rho(s, o_i) = P(o_i|s)$ and takes an action $a_i \in \mathcal{A}$ given by a policy, $\pi_i(o_i)$. Given actions from all the escorts, a joint action $a \in \textbf{A} \equiv \mathcal{A}^N$ is formed which induces transition in the environment according to the state transition function $\mathcal{T}(s, a, s') = P(s' | s, a)$. For the action $a_i$ in state $s$, the escort $i$ receives an individual reward $\mathcal{D}_i(s, a_i) \in D$ such that any action that improves the individual reward also improves the true global reward $R(s, a)$. Each escort individually tries to maximize their expected cumulative reward, $\displaystyle \mathop{\mathbb{E}}_{\tau \sim \pi} [D(\tau)]$, discounted by $\gamma$, where $\tau$ represents a sequence of states and actions of the escorts.

\section{Method}
\label{sec:methods}

The large continuous state space of the escorts motivates a deep RL  approximate solution for the Payload Protection Task.
While there exists many deep RL solutions, a class of Policy Gradient algorithms \cite{sutton-PG}, actor-critic methods \cite{Konda2003actor-critic} have been widely used in the RL scheme that train a value function, i.e., critic, using Bellman's equation to estimate the gradient of the performance. The gradient is then followed to update the policy, i.e., the actor. This reduces the variance thus stabilizing the training. Generalized Advantage Estimation (GAE) \cite{schulman-GAE} is an actor-critic method that improves sample efficiency and further stabilizes the learning by using an exponentially-weighted estimator of the advantage function as a baseline function and by using trust region optimization for both the actor and the critic.

We train multiple escorts that share a single GAE stochastic policy, an approach that is similar to Independent Actor-Critic with shared parameters \cite{foerster-COMA, gupta-coop-multi-agent}, using RLlib \cite{liang-RLlib}. The actor and critic are represented by two separate networks having the same architecture.
Sensor information must provide information about the payload, obstacles' and escorts' shape, location and velocity.  In order to obtain this information, we used simulated 1D LiDAR (512 outward rays, uniform spacing) from each escort. Object classification is implemented by concatenating three LiDAR distance measurements, one each to detect objects of a single type, i.e., escorts, payload and obstacles. For temporal reference that enables some inference of velocities, readings from the last three time steps are used.  This forms an array of size 3x1x1,536 (as shown in Fig.~\ref{fig:NN}). 

The output of the network is a set of actions for each escort that enables interception of obstacle threats.  This was implemented in the network by outputting a diagonal Gaussian distribution, $N([\mu_{V_x}; \mu_{V_y}], [\sigma_{V_x}; \sigma_{V_y}])$, where $\mu_{V_x}$ and $\mu_{V_y}$, and $\sigma_{V_x}$ and $\sigma{V_y}$ are the means and standard deviations of the escorts' horizontal and vertical speeds, respectively, from which continuous actions can be sampled.

The full network (Fig.~\ref{fig:NN}) encodes a policy that maps input sensor information to output actions.  We implemented this mapping through convolution layers (32 and 64 filters of size 1x10 and stride 1 with ReLU activation) each followed by a max pool layer (size 1x5 and stride 5). The output of the convolutional network is flattened and fed to a fully connected layer (size 512 with ReLU activation).

To train the escorts we design a reward function that acts as a signal to reinforce desired behavior. We parameterize this function with $\phi$ and define it as follows

\begin{equation}
\label{eqn:param_reward}
R_{\phi} = \phi^T[r_{goal}, r_{collision}, r_{cordon}, r_{step}]
\end{equation}
where it assigns $r_{goal}$ when the payload reaches the goal, $r_{collision}$ when an  collision occurs, $r_{cordon}$  when the cordon area is breached, and $r_{step}$ at every timestep.

\begin{figure}[htb!]
    \centering
    \begin{minipage}{0.9\linewidth}
        \centering
        \includegraphics[trim=0mm 0mm 0mm 0mm,clip,width=\textwidth]{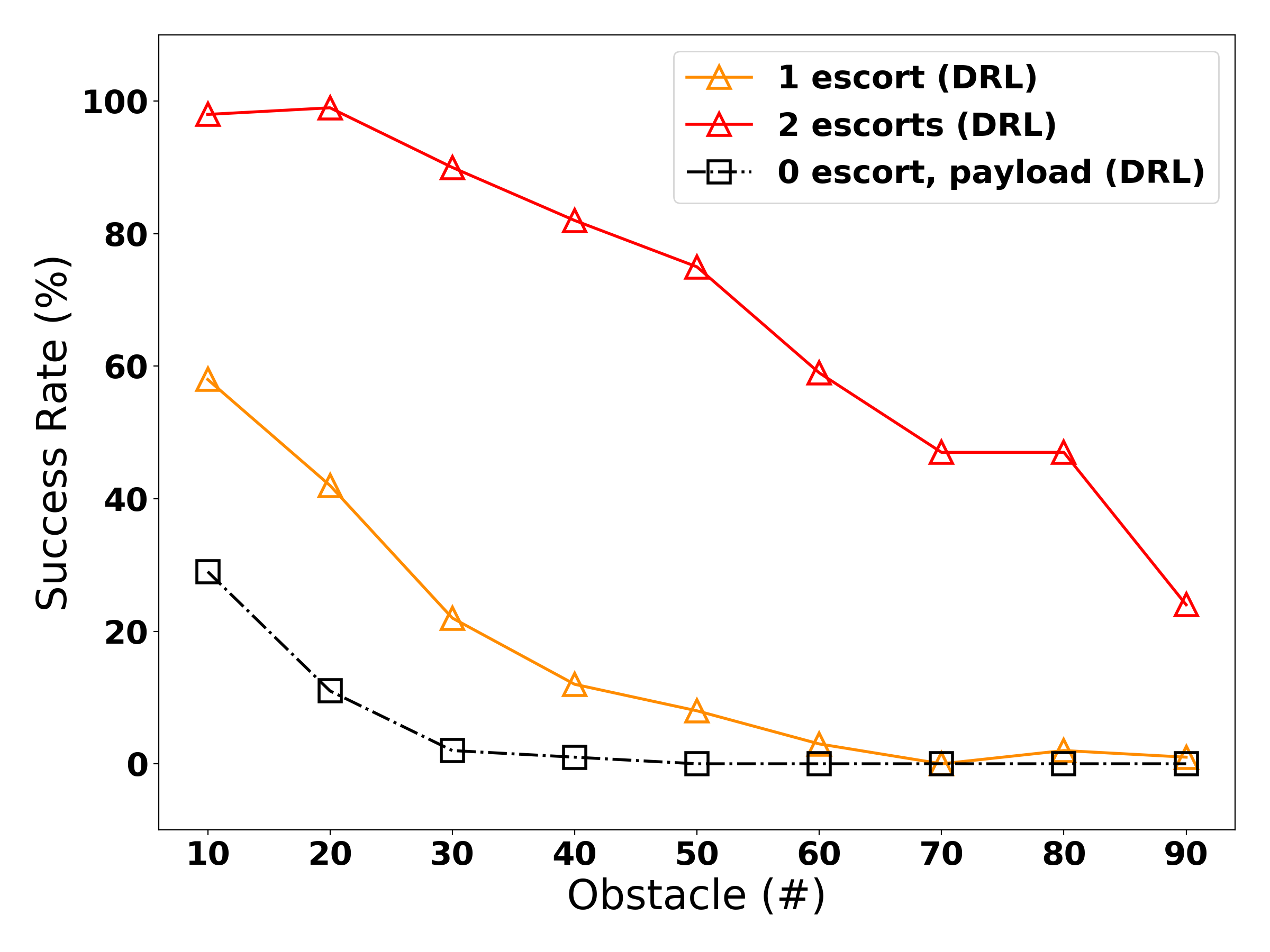}
        {\small (a)}
    \end{minipage}
    \begin{minipage}{0.9\linewidth}
        \centering
        \includegraphics[trim=0mm 0mm 0mm 0mm,clip,width=\textwidth]{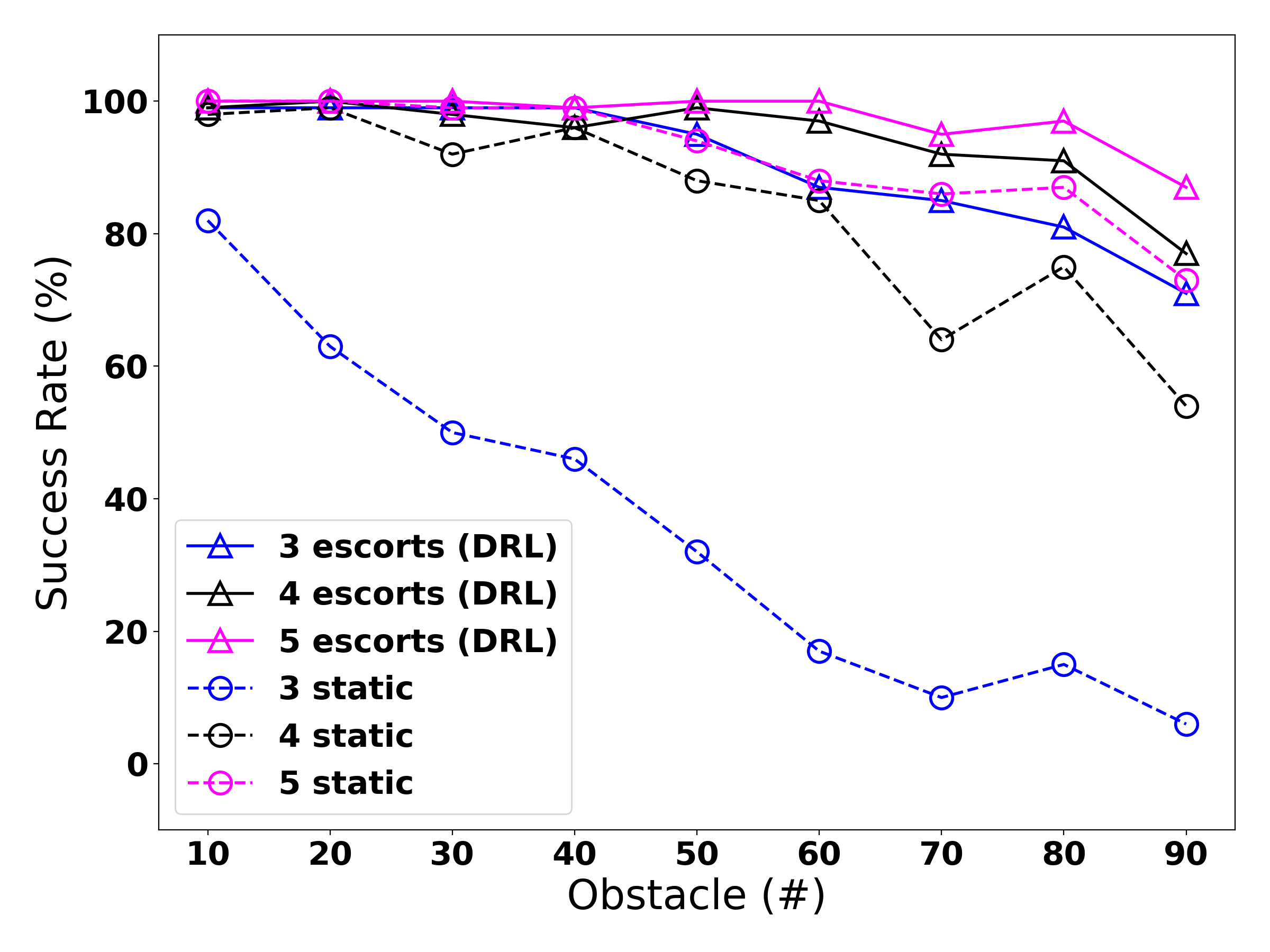}
        {\small (b)}
    \end{minipage}
    \begin{minipage}{0.9\linewidth}
        \centering
        \includegraphics[trim=0mm 0mm 0mm 0mm,clip,width=\textwidth]{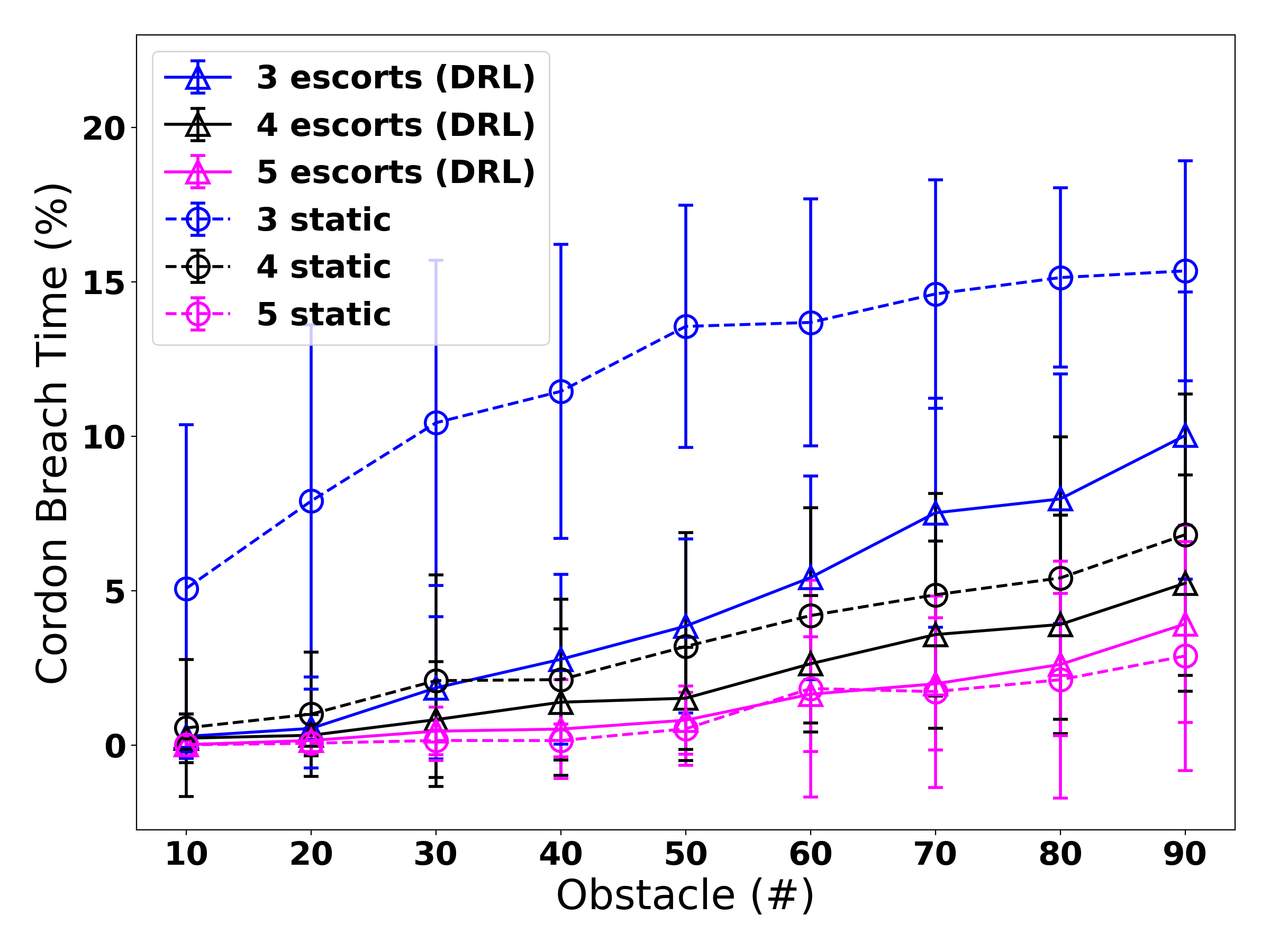}
        {\small (c)}
    \end{minipage}

    \caption{\scriptsize Demonstration of escort efficacy.  Navigation success rate in environments with increasing numbers of obstacles of (a) payloads not escorted but navigating with a deep RL obstacle avoidance policy (0 escort, payload (DRL)) and payloads navigating in a straight line and escorted with Dynamic deep RL escorts (deep RL)  and (b) payloads with escorts in a static formation (static) compared to deep RL escorts (DRL).  (c) Comparison of cordon area breach time for static vs. DRL escorts.  The number indicates the number of escorts.}
	\label{fig:goal_reached}
\end{figure}

\section{Experiments and Results}
In order to demonstrate the effectiveness of escorts, scenarios were set up to test the learned policy and the adaptability of trained escort teams.  We also compare the performance of escorts to some standard methods for enhancing the safety of navigation.  

\subsection{Experimental Setup}
For all  experiments the following specifications of the environment were used, unless specified otherwise. All objects are circular moving rigid bodies: a payload (radius 1.5m, maximum speed 0.25m/s), obstacles (radius 0.5m, maximum speed 1m/s), and holonomic escorts (radius 0.5m, maximum speed 3m/s).  
The dimensions of the environment are 50m by 50m as shown in Fig. \ref{fig:trainingEnvironment}. When the objects reach the boundary of the environment, they teleport and reappear at the opposite boundary. The radius of the cordon area is 5m.  The simulated LiDAR has a maximum vision distance of 8m, and escorts employ 512 beams at uniform intervals. At the beginning of each episode, obstacles are randomly assigned a position and a desired moving direction. Heuristics help facilitate setup by reducing states in collision and assisting the escorts to initially find the payload:  escorts are spawned within 2.5m to 3.5m from the payload, obstacles are at least 4.6m away from the payload, and the goal is exactly 20m away from the payload.

The obstacles interact both between themselves and with the escorts through the application of a social force model originally designed for interacting pedestrians \cite{helbing-SFM}.  In this model, temporal changes in preferred velocity, $\Vec{w_\alpha}$, of an obstacle $\alpha$, are defined as a summation, $\Vec{F}_\alpha(t)$, e.g., a \textit{social force} and typically normally distributed \textit{fluctuation}.  
\begin{align}
\label{eq:social_force}
\Vec{F}_\alpha(t) &:= \Vec{F_\alpha^0}
+ \sum_{\beta}\Vec{F}_{\alpha\beta}
+ \sum_{\gamma}\Vec{F}_{\alpha\gamma}
, \\
\dfrac{d\Vec{w_\alpha}}{dt} &:= \Vec{F}_\alpha(t) + fluctuation
\end{align}
The \textit{social force} takes into account the tendency of an obstacle to reach a certain desired velocity with a relaxation time of 0.7s, given by $\Vec{F_\alpha^0}$, and the \textit{repulsive effects} of other obstacles, $\beta$, and escorts, $\gamma$, given by $\Vec{F}_{\alpha\beta}$ and $\Vec{F}_{\alpha\gamma}$, respectively in Eq. \ref{eq:social_force}.
The repulsive potential of amplitude 7.8$m^2/s^2$ is assumed to decrease exponentially in the form of an ellipse that is directed into the direction of motion. The repulsive effects are only felt if the objects are within the influential radius (5m) and inside the directionally dependant vision cone  of the obstacle (200$\degree$).
In our setup, the escorts can apply social forces on the obstacles.  However, the payload does not apply social forces.  This is due to the assumption that the payload is of high consequence and is similar to other work with distracted  pedestrians \cite{kayukawa-bbeep}.

The Payload Protection Task requires both collision free navigation and protection of the cordon area. As such, both metrics are used to assess the success of the learned policy.  Success rate directly measures ability of the policy to navigate without collision. It does so by providing the ratio of runs where the payload is able to  navigate to the goal without collision over all runs. Another metric, cordon area breach time, captures the amount of time that obstacles are in the cordon area. It is averaged over all runs and is presented as percentage of the time the cordon area is breached over the total duration of the episode. All experiments are averaged over 100 iterations unless otherwise specified. We employed GNU Parallel \cite{tange-parallel} to evaluate experiments in parallel.

\subsection{Escort Policy Training}
Since the task is fully cooperative and our escorts are homogeneous in terms of their goals and capabilities, we reward all the escorts by the same global reward, $R(s, a)$, that is a function of the current state $s$ and the joint action $a$ of all the escorts.
At each time step, we compute the reward function, given by Eq. \ref{eqn:param_reward}, where we set $\phi$ to a unit vector, $r_{goal}$ to 1, $r_{collision}$ to -1, $r_{step}$ to 0.01 and $r_{cordon}$ to a function that penalizes for every obstacle that penetrates the cordon area proportional to their proximity to the payload and is defined as follows
\begin{equation}
\label{eqn:breach_reward}
r_{cordon} = -c \smashoperator{\sum_{\{{o_i} | d({\bm{x}^p}_{t}, {\bm{x}^{o_i}}_{t}) < S_{cordon}\}}} (1 - \frac{d({\bm{x}^p}_{t}, {\bm{x}^{o_i}}_{t})}{S_{cordon}}),
\end{equation}
where ${\bm{x}^{o_i}}_{t}$ and ${\bm{x}^p}_{t}$ are the positions of the obstacle, $o_i$, and the payload, $p$, at time step $t$, respectively, $d({\bm{x}^p}_{t}, {\bm{x}^{o_i}}_{t})$ is the distance between them, $S_{cordon}$ is the radius of the cordon area and $c$ is a constant (0.05 for this work). We define collision if there is a collision between objects of any two different types, i.e., \textit{payload-obstacle}, \textit{payload-escort} and \textit{escort-obstacle}. We terminate the episode if there is a collision or if the goal is reached.

We use a single GAE policy that is shared between all the escorts. To train this policy we collect experience samples in parallel on 100 cores of Intel Xeon E-2146G @ 3.50 GHz. We train the policy every time a training batch of size 524,288 samples is collected by performing stochastic gradient descents of mini-batch size of 65,536 samples on 4 NVIDIA Tesla V100 GPUs in parallel. We use mean of the rewards of all the samples in a training batch as a metric for convergence that typically occurs in 100M samples and takes about 24hrs.

\excise{
\begin{figure}
    \centering
    \includegraphics[scale=0.28]{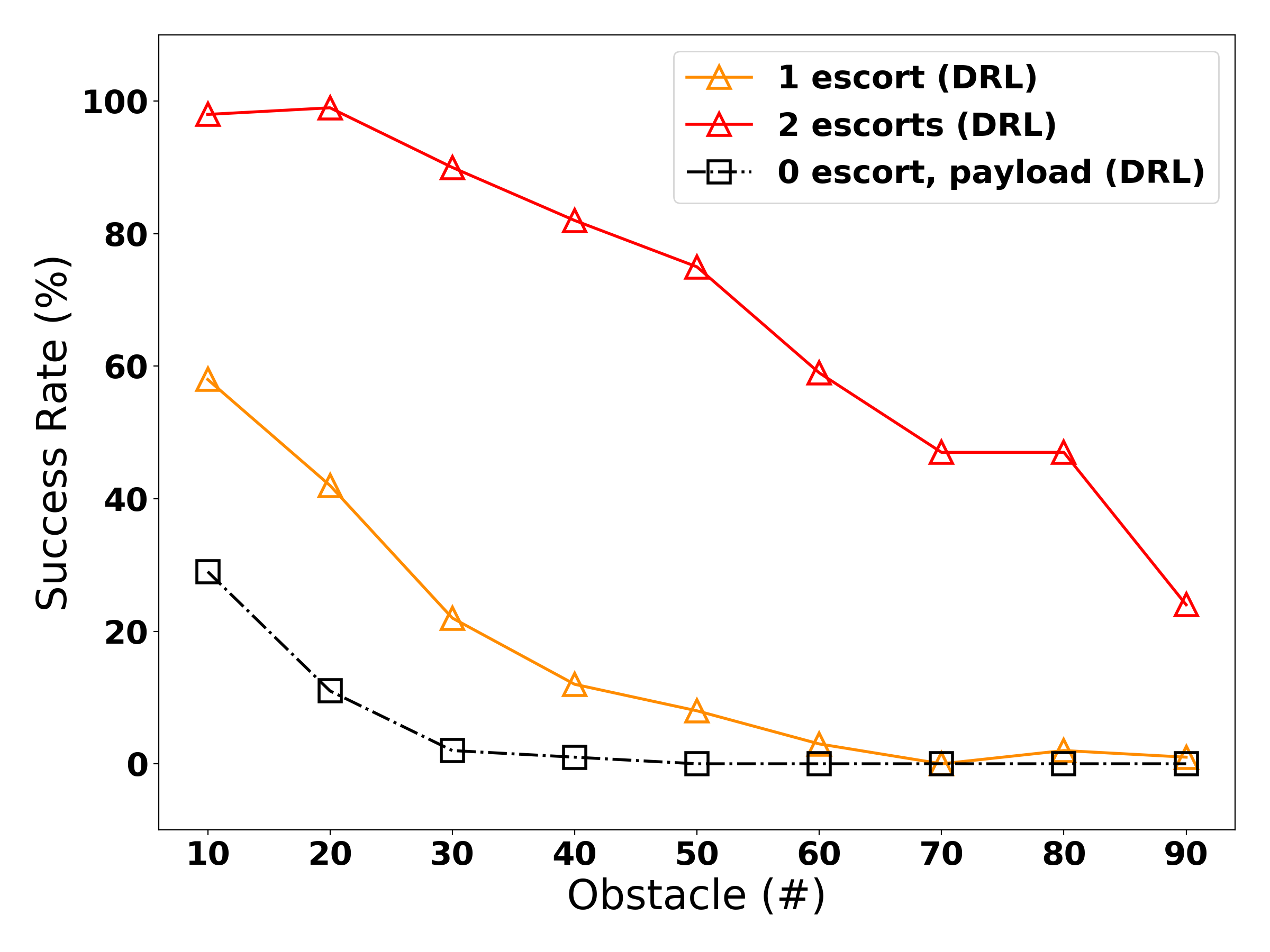}
        \caption{\scriptsize Success rate, defined as reaching the goal without collision, while navigating with an increasing number of dynamic obstacles.  The black line, {\em 0 escorts, payload (DRL)}, demonstrates navigating with the payload using a deep RL solution trained solution for avoiding moving obstacles.  The orange and red lines show a payload navigating in a straight line to the goal with 1 (orange) or 2 (red) deep RL escorts.}
	\label{fig:three}
\end{figure}
}

\excise{
\begin{figure}
    \centering
    \includegraphics[scale=0.28]{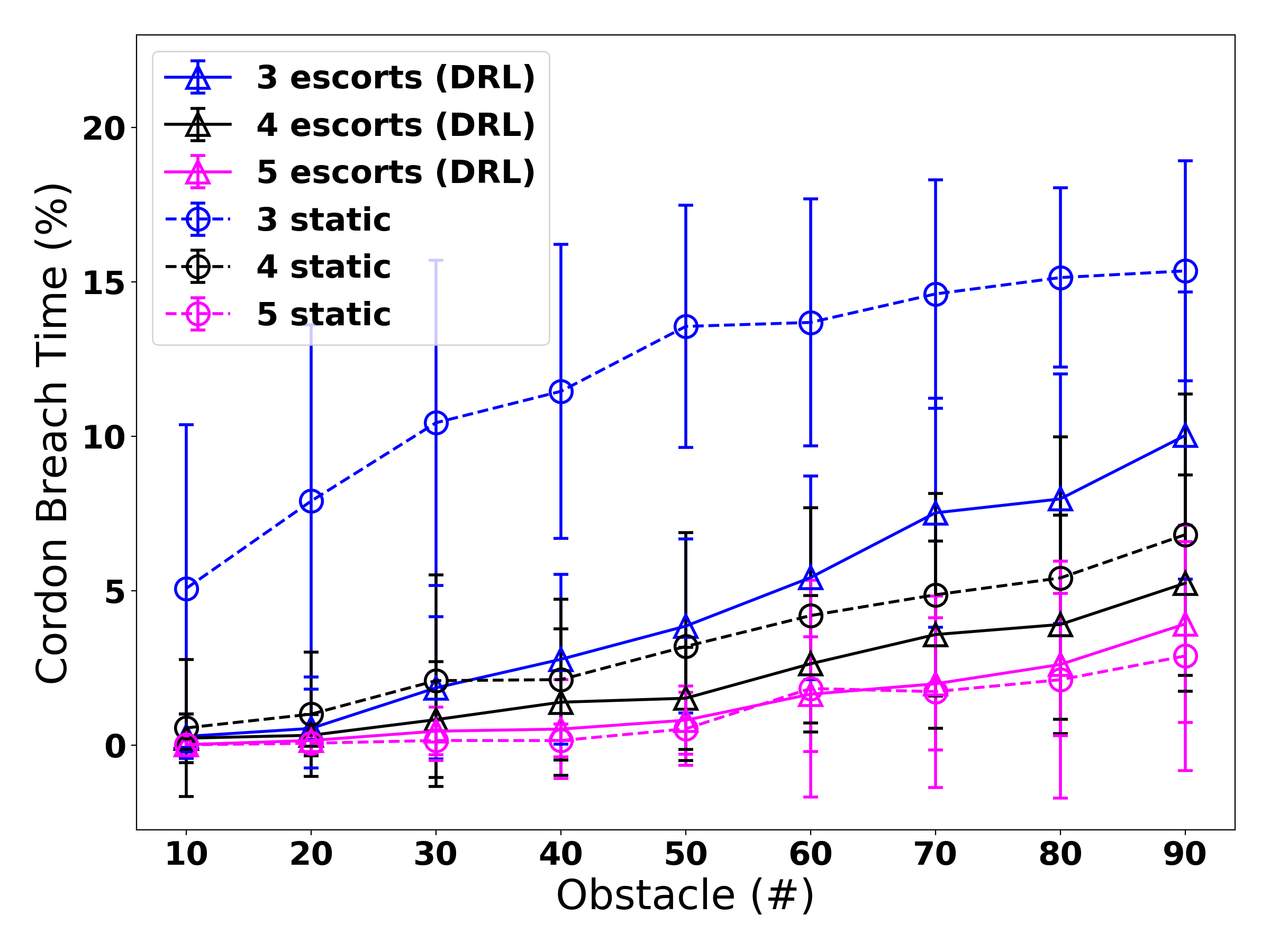}
    \caption{\scriptsize Cordon area breach time, the amount of time an obstacle is in the cordon area of the payload during payload navigation, with Dynamic deep RL escorts (deep RL) compared to static escorts (static) uniformly placed around the payload.  The number indicates the number of escorts.}
    \label{fig:tb_timesteps}
\end{figure}
}

\subsection{Escort Efficacy}

To demonstrate the impact of autonomous escorts, we evaluated trained escorts against two conventional approaches for enhancing navigation safety (Fig.~\ref{fig:goal_reached}).  In the first comparison, we evaluated a limited number of escorts against an approach where the payload was trained to adjust its navigation dynamically in order to avoid moving obstacles.  This deep RL navigation solution was presented as an approach where the learned value function could, from observed obstacle motions, predict probabilities of collision \cite{arpit-RLvsSR}. However, the predictions are challenging, as the obstacles are interacting. This increased difficulty is seen in Fig.~\ref{fig:goal_reached}(a) where the success rate plummets to 2\% when the number of obstacles in the scenario is above 20. However, even a single escort increases the success rate by 31\% at the 30 obstacle scenario. The impact of escorts is more apparent with a team of two escorts.  The team finds a solution 24\% of the time in the most challenging scenario (90 obstacles) and also demonstrates up to 68\% increased success over single escort runs. 
\excise{
In the scenario where the payload itself can exert social forces onto the obstacles around it, the success rate is comparable to having 3 escorts with a payload that exerts no forces.
}
In a second comparison, we demonstrate the impact of autonomous escorts compared to escorts placed in a fixed formation in a uniform distribution around the payload. This static formation is intended to form a `protective barrier' keeping obstacles away from the payload and out of the cordon area. Intuitively, the number of escorts in the team impacts the amount of protection for both static and deep RL trained escorts, as seen in Fig.~\ref{fig:goal_reached}(b); increasing the number of escorts increases the success rate of the navigation. However, the performance of 3 deep RL escorts is comparable to having 5 static escorts. As such we can achieve a desired degree of safety with fewer escorts if we use deep RL.
Additionally, trained deep RL escorts are able to enhance protection of the cordon area, as seen in Fig.~\ref{fig:goal_reached}(c).  While this impact is diminished as the static escorts saturate the region around the payload, deep RL escorts provide an improvement over static ones below saturation points. We also consider a scenario where the payload itself can exert social forces onto the obstacles around it. In this scenario, the success rate is nearly as good as to having 3 deep RL escorts protect a payload that exerts no forces (results not shown).

\subsection{Robustness to Noise}
In an extension to the base problem, we also explore how trained escort teams adapt when there is a disruption or disturbance in the scenario. These issues can occur for many different reasons. However, we look at two practical disruptions: an unexpected change in the reaction of the obstacles to an expected social force outcome, and a change in the payload, i.e., a change in size that represents a reconfiguration of the payload.  As these disruptions may be difficult to account for during training, the results shown are produced with no retraining. 

\begin{figure}[htb!]
    \centering
    \begin{minipage}{0.9\linewidth}
        \centering
        \includegraphics[trim=0mm 0mm 0mm 0mm,clip,width=\textwidth]{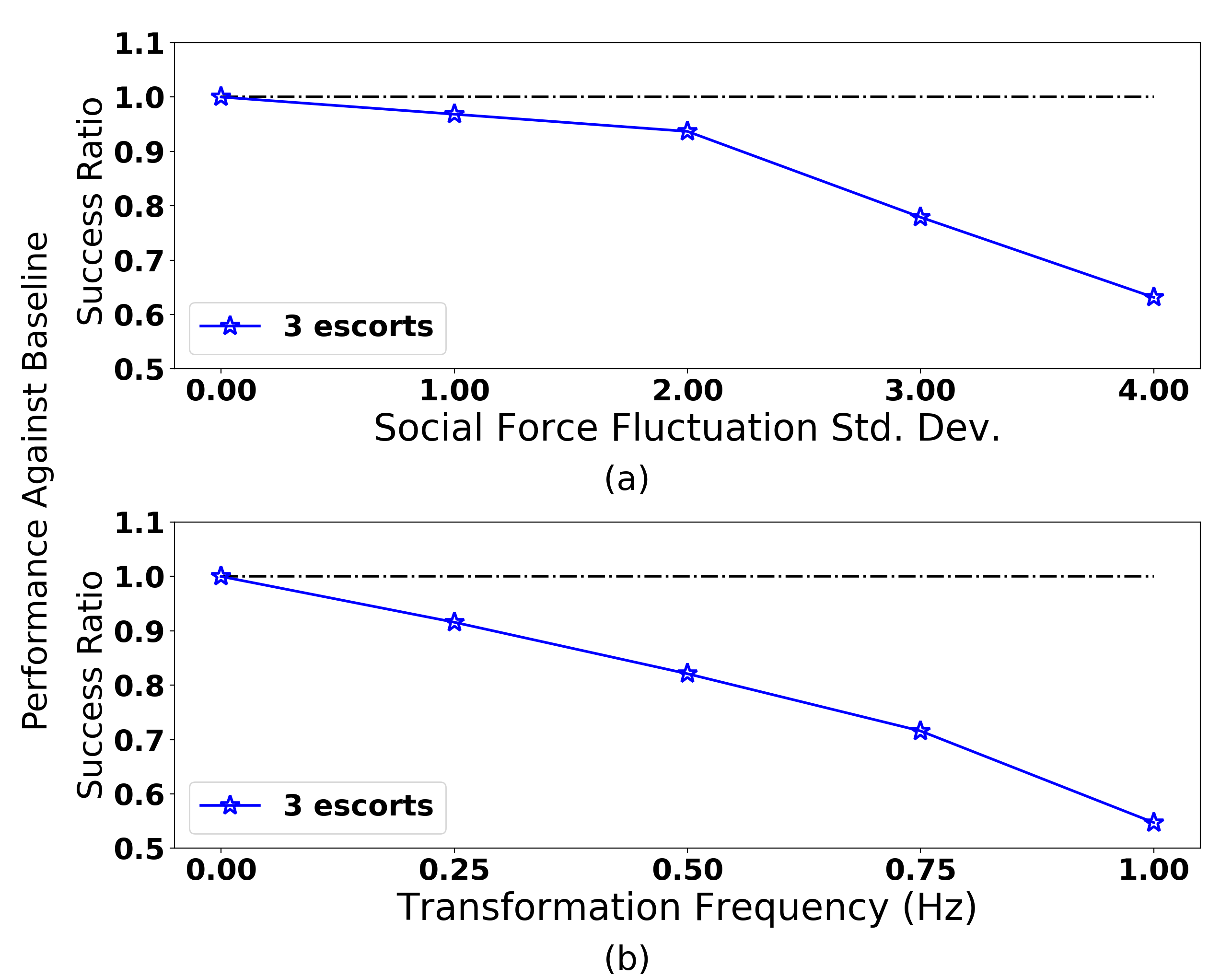}
    \end{minipage}

    \caption{\scriptsize Success rate expressed as a ratio to a baseline problem of three deep RL escorts and 50 obstacles for disruptions: (a) increased standard deviation of $fluctuations$ in the social forces and (b) increased rate of transformation of the payload size.}
	\label{fig:additionalChanges}	\label{fig:sf_noise}

\end{figure}


Fig.~\ref{fig:sf_noise}(a) shows the impact of unexpected disruptions in the social force model compared to the baseline case without disruptions, in an environment with 3 deep RL escorts and 50 obstacles. For this scenario, three escorts are trained without any disruption in the social force model, $fluctuation=0$.  However, post-training, the social force $fluctuation$ term becomes non-zero, thus making the social force interactions increasingly unpredictable. As expected, collision-free payload navigation success drops as the social force fluctuations increase.  However, even under moderate noise with standard deviation of 2.0, the escorts successfully defend the payload 93\% of the times that of the baseline with no fluctuation.



\begin{figure}[htb!]
    \centering
    \begin{minipage}{0.7\linewidth}
        \centering
        \includegraphics[trim=0mm 0mm 0mm 0mm,clip,width=\textwidth]{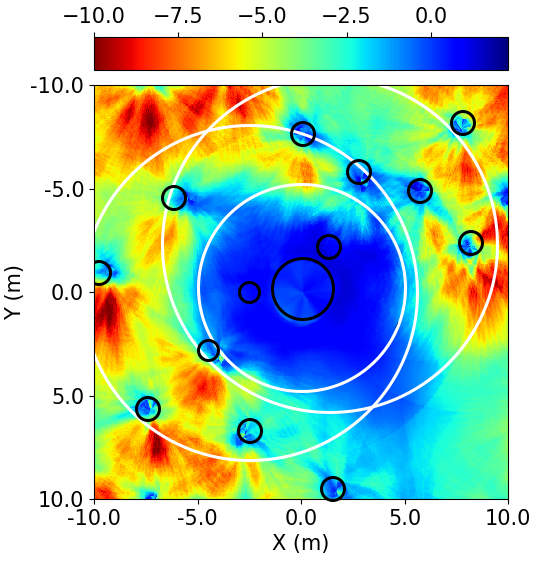}
        {\small (a)}
    \end{minipage}
    \begin{minipage}{0.7\linewidth}
        \centering
 \includegraphics[trim=0mm 0mm 0mm 0mm,clip,width=\textwidth]{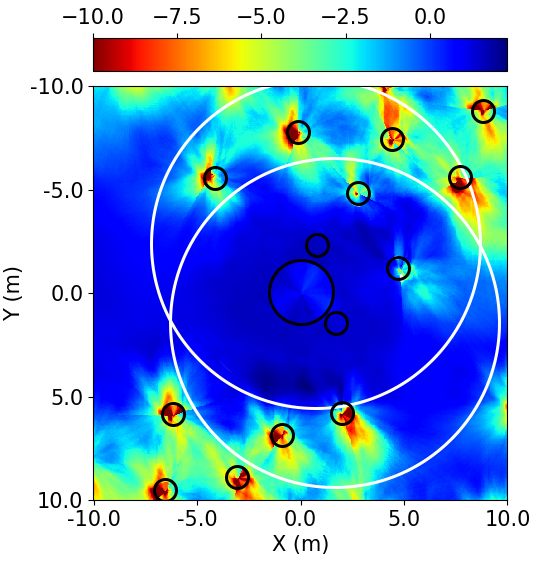}
        {\small (b)}
    \end{minipage}
    \begin{minipage}{.9\linewidth}
        \centering
               \includegraphics[trim=0mm 0mm 0mm 0mm,clip,width=\textwidth]{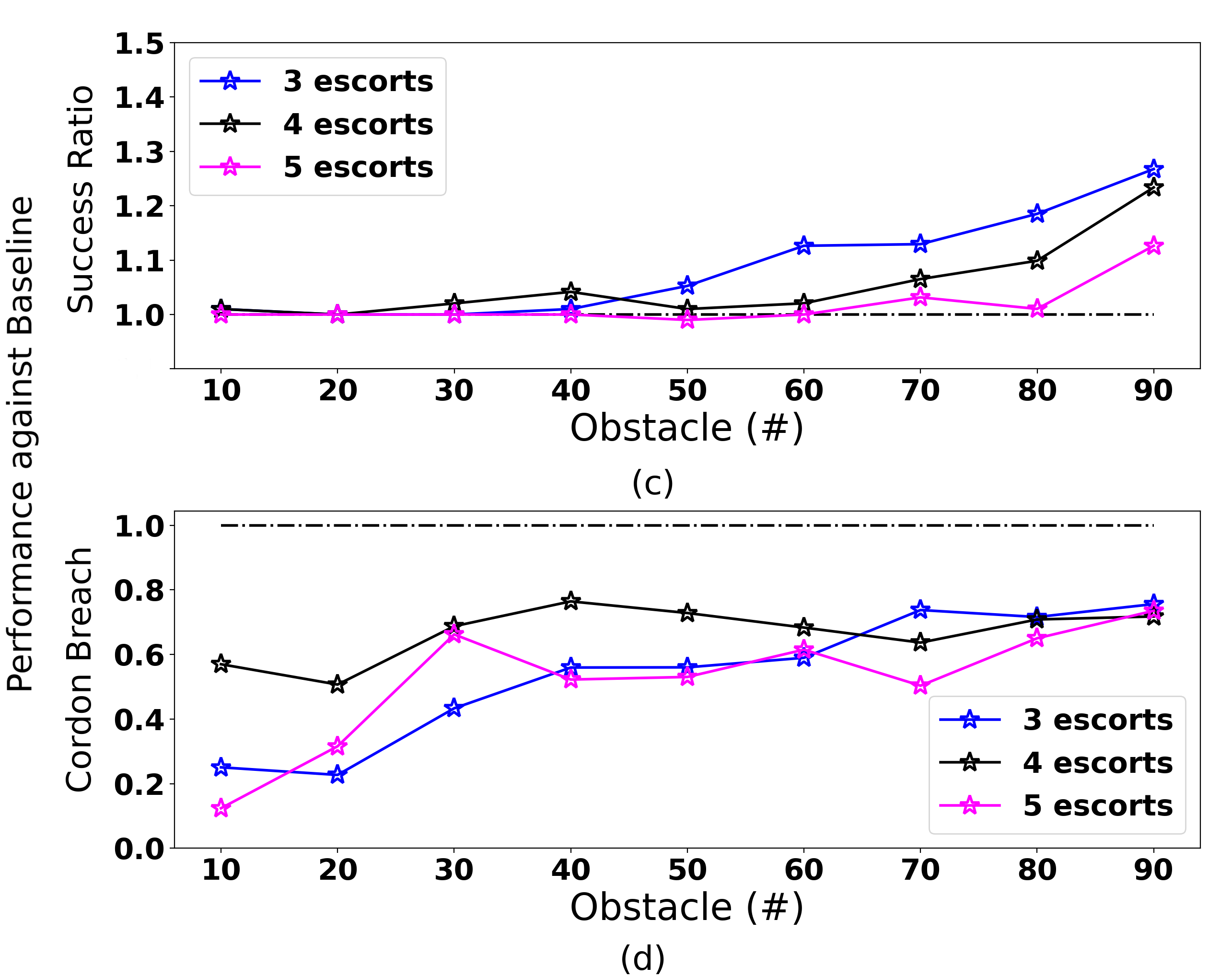}
    \end{minipage}

    \caption{\scriptsize (a-b) Value functions for the trained escorts for (a) the Payload Protection Task and (b) Payload Protection Task without penalty for obstacles entering the cordon area.  Black circles represent  payload, escorts, and obstacles.  White circles represent cordon area (where applicable) and sensor ranges of escorts.  (c-d) Performance of retrained policy without consideration of a cordon area penalty compared to a baseline with the same number of escorts and policy considering the cordon area expressed in both (c) success rate and (d) cordon area breach time.  }
	\label{fig:reward_PayloadOnly}
\end{figure}


In Fig.~\ref{fig:sf_noise}(b) we explore the adaptability of the learning policy to a change in the payload.  In practice, this change could be a reconfiguration of a multi-body payload to a new conformation.  In order to test this, we implemented a payload of increasing and shrinking radius.  Frequency represents a change in state over time, i.e., a transformation frequency of 0.5 represents a change to a radius of 2.5m from the original 1.5m in 2 seconds. Results with transformations were tested in an environment with 50 obstacles and three deep RL escorts.  While the disruptions demonstrate a linear loss in success rate, as compared to the same setup without the transforming payload, the escorts show great efficacy still defending the payload while it is navigating. At a frequency as high as 1 Hz, the escorts are still more than 50\% as successful at protecting the payload as they are at protecting a non changing one.

\subsection{Changing Policies}

Some scenario changes may require retraining in order to enhance the performance of the escort team to address the modified task.  In this section, we explore these cases by looking at changes in the policy that enable the escorts to solve a similar, yet different task.

For the first scenario, we change the reward function, eliminating the penalty for breaching the cordon area (Eq.~\ref{eqn:breach_reward}).  This modifies the task to one where the escorts are strictly interested in protecting the payload. After training with this new reward function, we can see that escorts do a better job protecting the payload from obstacles, thus achieving a higher success rate by being up to 26\% better at navigating the payload to the goal compared to the baseline (Fig. \ref{fig:reward_PayloadOnly}(c)). However, this is at the cost of protecting the cordon area where the escorts perform only up to 12\% as good (Fig. \ref{fig:reward_PayloadOnly}(d)). We can visualize the change in policy by plotting the value functions for both policies. To do so, for each point around the payload (10m by 10m) we take observations over 3 timesteps to form a state. Using this state we query the critic net to produce the state-value. In Fig. \ref{fig:reward_PayloadOnly}(a-b), we generate the heat maps for escorts trained for the standard Payload Protection Task (Fig.~\ref{fig:reward_PayloadOnly}(a)), and the modified Payload Protection Task without considering cordon area breach (Fig.~\ref{fig:reward_PayloadOnly}(b)). We can observe that the standard escorts prefer to maintain the cordon area, despite not being told explicitly what the size of the cordon area was, whereas the modified escorts have no strong preference for maintaining the cordon area. We can also see that the standard escorts are more concerned with the obstacles and try to avoid steering them into the cordon area. On the other hand, the modified escorts are not as concerned by influencing the obstacles in the wrong direction, and care more about steering them away from the payload without colliding with them.

\begin{figure}
    \centering
    \includegraphics[scale=0.23]{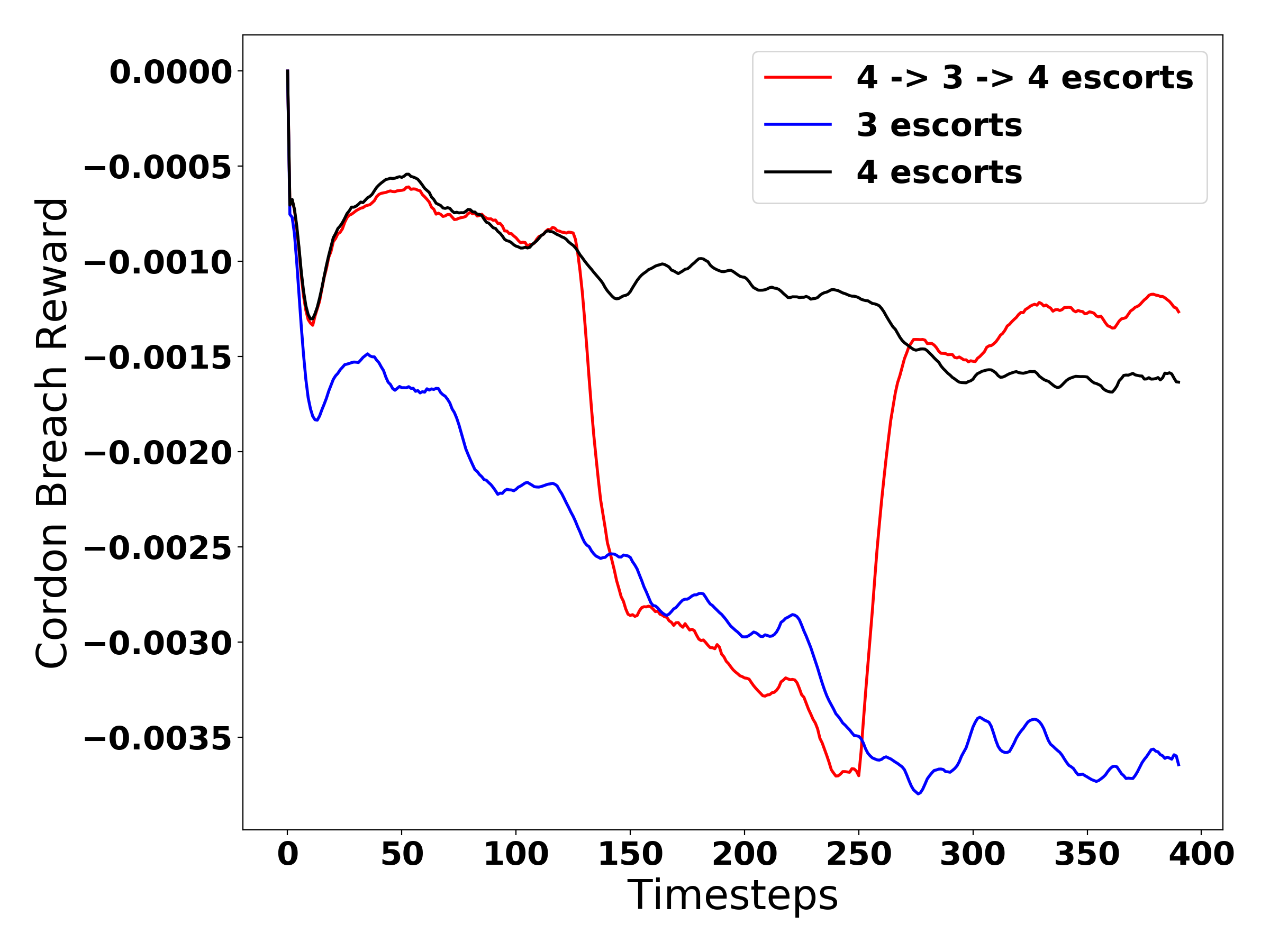}
        \caption{\scriptsize Average cordon area breach reward over the course of successful payload navigation runs in an environment with 90 obstacles.  Lines show a case with 3 deep RL escorts, with 4 deep RL escorts, and that starts with four escorts, drops to three (at timestep 125), and then increases back to four (at timestep 250).}
	\label{fig:escortLossGain}
\end{figure}

The deep RL escorts can also be trained to be robust to losses and gains in escorts. For this scenario, we trained the escorts with a variable number of escorts, from one to six, and then post learning dynamically added and removed escorts while the payload is navigating. Since the network structure was robust to to this modification, the only difference in training was the existence of varying numbers of escorts in each escort's field of view. Training on varying numbers of escorts facilities online adaptation to gains and losses of escorts. The adaptation of escorts to these sudden losses are best seen in the rewards from successful runs, as shown in Fig.~\ref{fig:escortLossGain}.  As a reminder, the cordon area reward penalizes for every obstacle that breaches the cordon area proportional to their proximity to the payload (Eq. \ref{eqn:breach_reward}).  For a baseline comparison, the cordon area breach reward curves for three and four deep RL escorts are plotted. These are compared to rewards when the escorts are dynamically changing over the run from 4 to 3 back to 4 escorts. While it takes about 15 timesteps (3 sec) to reposition the escorts, they are able to adjust to changes in the number of escorts. For all runs, it takes about 56 timesteps (11.2 sec) for escorts to initially position themselves from the spawned positions to the maximum reward. Lines in Fig.~\ref{fig:escortLossGain} were averaged over 1000 runs for plot clarity.



\section{Conclusion}
Defensive escorts help provide critical safety for high-value payloads that are navigating.  Escorts work by coordinating their actions in order to protect the payload and can also be trained to provide a safe cordon area around the navigating payload.  They enhance safety over what current solutions for obstacle avoidance can provide in crowded environments, and can be robust to several changes including disruptions in the system, changes in payload size, and gain and loss of escorts.  Our deep RL solution provides an end-to-end solution for escort coordination  toward the team goal of payload protection.  With only partial observations of the environment and no other explicit information, the escorts learn to automatically coordinate their positions.  Additionally, we demonstrate modifications to the Payload Protection Task that modify the learned value function, and change the goals of the escorts.  

\section{Acknowledgements}
 We especially would like to thank Lewis Chiang for his early help with our literature review for defensive escorts and his creative discussions.  We would also like to thank Evan Carter of the Army Research Lab for helpful suggestions and discussions on this work. IP protected by U.S. Provisional Patent (STC.UNM.EDU).


\bibliography{IEEEabrv,temp.bib}

\end{document}

%% file: macros.tex
\usepackage{graphicx} 
\usepackage{epstopdf}
\DeclareGraphicsExtensions{.eps}

\usepackage{multirow}
\usepackage{makecell}
\usepackage{tabularx}
\usepackage{algorithmic}
\usepackage{algorithm}
\usepackage{amsmath} 
\usepackage{amssymb}
\usepackage{amsxtra}
\usepackage{url} 
\usepackage{color} 
\usepackage{bm}
\usepackage{placeins}
\usepackage[caption=false]{subfig}
\usepackage{xspace}
\usepackage{rotating}
\usepackage{siunitx}
\usepackage{textcomp}%

\newcommand{\excise}[1]{}

\newif\ifremark
\long\def\remark#1{
  \ifremark%
  \begingroup%
  \dimen0=\textwidth
  \advance\dimen0 by -1in%
  \setbox0=\hbox{\parbox[b]{\dimen0}{\protect\em #1}}
  \dimen1=\ht0\advance\dimen1 by 2pt%
  \dimen2=\dp0\advance\dimen2 by 2pt%
  \vskip 0.25pt%
  \hbox to \textwidth{%
    \vrule height\dimen1 width 3pt depth\dimen2%
    \hss\copy0\hss%
    \vrule height\dimen1 width 3pt depth\dimen2%
  }%
  \endgroup%
  \fi}

\newcommand{\ui}[1]{a}

